\documentclass[lettersize,journal]{IEEEtran}
\usepackage{amsmath,amsfonts}
\usepackage{algorithmic}
\usepackage{algorithm}
\usepackage{array}
\usepackage[caption=false,font=normalsize,labelfont=sf,textfont=sf]{subfig}
\usepackage{textcomp}
\usepackage{stfloats}
\usepackage{url}
\usepackage{verbatim}
\usepackage{graphicx}

\usepackage{multirow} 
\usepackage{soul}
\usepackage[utf8]{inputenc}
\usepackage{graphicx}
\usepackage{amsmath}
\usepackage{amsthm}
\usepackage{booktabs}
\usepackage{algorithm}
\usepackage{algorithmic}
\usepackage[switch]{lineno}
\usepackage{xcolor}
\usepackage[table]{xcolor}
\definecolor{lightgray}{gray}{0.88}

\usepackage{cite}
\begin{document}

\title{A Robust Point Cloud Analysis Framework Inspired By Primary Visual Cortex}

\author{Jisheng Dang, Dengyue Pan, Delin Deng, Yifan Zhang, Bimei Wang,  Hong Peng, Bin Hu,  ~\IEEEmembership{Fellow, ~IEEE}, Qi Tian,  ~\IEEEmembership{Fellow, ~IEEE},   Tat-Seng Chua

\thanks{This work was supported by the National Natural Science Foundation of China (Grant No. U24B20186). This work was also supported by the Supercomputing Center of Lanzhou University.}

\thanks{Jisheng Dang, Dengyue Pan, Delin Deng, Yifan Zhang, Bimei Wang, and Hong Peng are with the School of Information Science and Engineering, Lanzhou University.  Jisheng Dang and Dengyue Pan contributed equally to this work. (E-mail: wangbm@stu2021.jnu.edu.cn, dangjisheng@lzu.edu.cn, pengh@lzu.edu.cn).}  


\thanks{Bin Hu is with the School of Medical Technology, Beijing Institute of Technology, Beijing 100081, China (E-mail: bh@bit.edu.cn). }

\thanks{Qi Tian is with Cloud and AI BU, Huawei, Shenzhen, Guangdong 518129,
China (e-mail: tian.qi1@huawei.com).}

\thanks{Tat-Seng Chua is with the School of Computing, National University of Singapore, Singapore 119077 (E-mail: dcscts@nus.edu.sg).}

\thanks{*Corresponding author: Hong Peng, Bin Hu,  and Bimei Wang.}

 }

\maketitle

\begin{abstract}
Despite significant advancements in point cloud analysis, reducing energy consumption and improving robustness remain understudied, largely due to the inherent limitations of Convolutional Neural Networks (CNNs). To address this, we draw inspiration from the primary visual cortex and propose a Dendritic-Connected Continuous-Coupled Neural Network (DC-CCNN), a novel Brain-Inspired Neural Network (BINN) architecture for point cloud analysis. By combining discrete and continuous encoding, our design replaces traditional Multilayer Perceptrons (MLPs) with more efficient and robust BINNs. Building upon this framework, we further propose an extended model, \textbf{DC-CCNN++}, to improve robustness under complex corruption conditions. Specifically, we introduce a \textbf{Neuro-Inspired Robust Modulation-and-Readout Module (NRMR)} to enhance feature stability and decision robustness through global-context gain modulation and dual-code evidence integration, and we design a \textbf{Cortically Inspired Progressive Variability Training (CPVT)} strategy, which progressively exposes the model to structured environmental variability while preserving stable clean-sample anchors during training. Experimental results show that DC-CCNN++ improves the performance of brain-inspired networks on point cloud analysis while maintaining performance comparable to state-of-the-art methods. Compared with the original DC-CCNN, it achieves stronger results on both classification and part segmentation, and exhibits enhanced robustness against sparsity, occlusion, Gaussian noise, salt-and-pepper noise, and spatial transformations. With its efficiency, robustness, and biologically grounded design, DC-CCNN++ provides a promising alternative to traditional deep learning methods for point cloud analysis. Codes are available at \url{https://anonymous.4open.science/r/DC-CCNNpp-44E3}.
\end{abstract}


\section{Introduction}
Point cloud analysis plays a central role in 3D understanding, drawing increasing attention from both academia and industry~\cite{zheng2025pad}. Unlike 2D images, point clouds consist of unordered, irregularly spaced 3D points, making conventional image processing ineffective. Sparsity and noise further increase the difficulty. To address these challenges, various approaches have been explored, including geometric methods, deep learning, and Graph Neural Networks (GNNs).
\begin{figure}[t!]
\centering
\includegraphics[height=5.7cm]{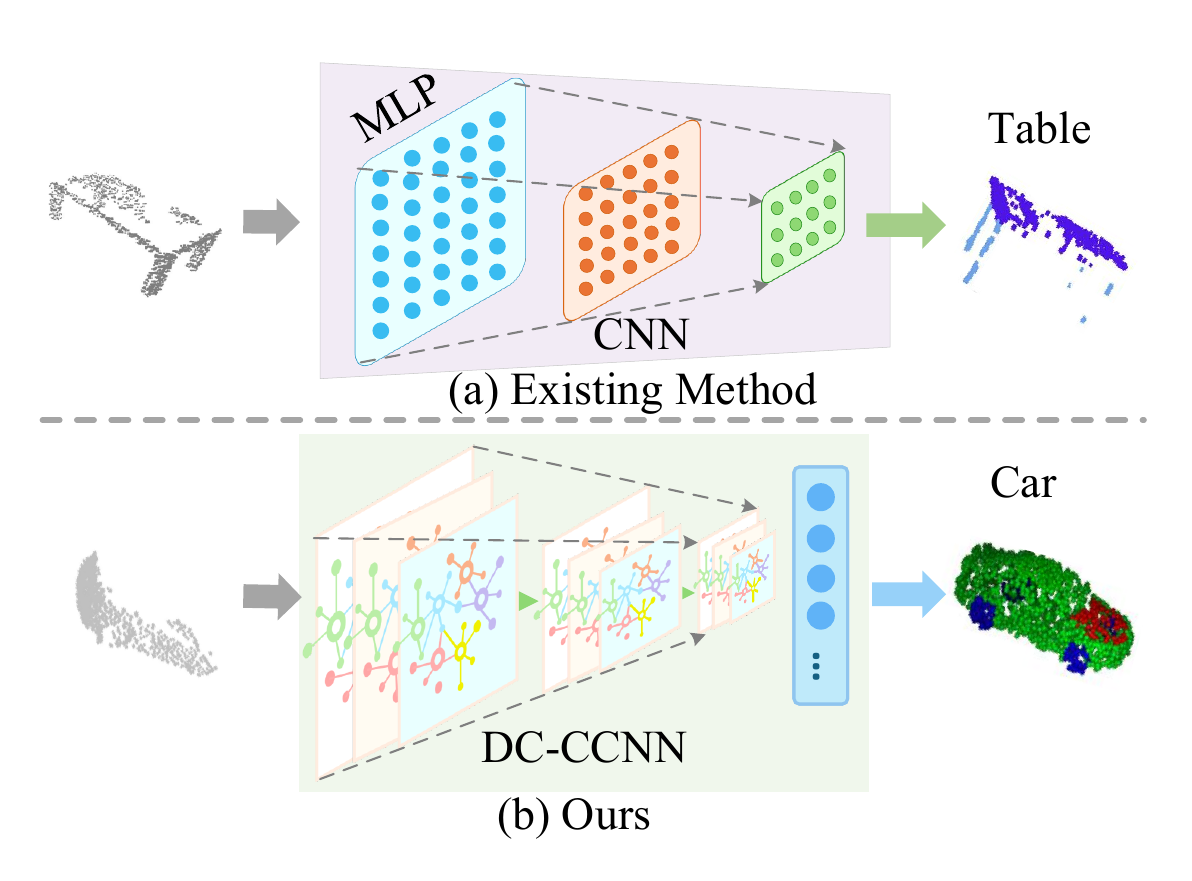}\\
\caption{Comparison of CNN-based methods and our DC-CCNN for 3D point cloud classification. DC-CCNN achieves similar accuracy with significantly lower energy consumption.}
\label{fig:intro1}
\end{figure}
Deep learning methods for point clouds are generally categorized by their data representation strategies. Indirect methods project point clouds onto grids (e.g., voxels~\cite{lei2024densetrack} or images~\cite{dang2023unified,danglhpgcnn,dangadaptive,dangtemporo}) to reuse 2D CNNs, but this often causes information loss. Direct methods, such as PointNet~\cite{qi2017pointnet}, PointNet++~\cite{qi2017pointnet++}, and PointMLP~\cite{ma2022rethinking}, process raw point sets using shared MLPs, improving efficiency and accuracy. However, most models assume clean data and degrade under corruption~\cite{dong2023benchmarking}. Robust 3D classification remains an open challenge.

Despite progress, efficiently handling large-scale, sparse, and noisy point clouds remains difficult. CNN-based methods often suffer from high computational cost, limited robustness to deformation, and poor adaptability to irregular structures~\cite{dang2020hpgcnn,9413879,dang2024beyond}. Their reliance on dense convolutions restricts deployment on low-power devices and hampers generalization under occlusion or noise.

To address these limitations, in our earlier work we proposed \textit{Dendritic-Connected Continuous-Coupled Neural Network (DC-CCNN)}, a novel Brain-Inspired Neural Network (BINN) architecture designed for robust and efficient point cloud analysis. As illustrated in Figure~\ref{fig:intro1}, DC-CCNN leverages Spiking Neural Networks (SNNs)~\cite{wu2025neural}, incorporating biological sensitivity to spatial and temporal patterns. Through spike coding and event-driven computation, it reduces complexity while improving robustness to deformation, occlusion, and noise. The integration of discrete and continuous dynamics enables high efficiency with low energy consumption, achieving performance competitive with state-of-the-art models using fewer computational resources.

Building upon this framework, in this paper we further propose an extended model, \textit{DC-CCNN++}, to improve robustness and generalization under complex corruption conditions. Specifically, we introduce a \textit{Neuro-Inspired Robust Modulation-and-Readout Module (NRMR)}, which enhances the original DC-CCNN with global-context gain modulation during feature formation and dual-code evidence integration during classification, thereby improving feature stability and decision robustness. In addition, inspired by cortical development under fluctuating sensory stimuli, we design a \textit{Cortically Inspired Progressive Variability Training (CPVT)} strategy, which progressively exposes the model to structured environmental variability while preserving stable clean-sample anchors during training. In this way, DC-CCNN++ extends the brain-inspired robustness of the original DC-CCNN from both the architectural and training perspectives.

DC-CCNN++ offers strong potential for applications in 3D vision, robotics, and autonomous driving. The main contributions of this work are:
\begin{itemize}
    \item We propose DC-CCNN++, an extended brain-inspired architecture built upon DC-CCNN, which further improves robustness and generalization under complex corruption conditions while preserving the advantages of discrete--continuous collaborative coding.
    \item We introduce a Neuro-Inspired Robust Modulation-and-Readout Module  to enhance representation stability and decision robustness through global-context gain modulation and dual-code robust readout.
    \item We design a Cortically Inspired Progressive Variability Training  strategy, which improves adaptation to sparsity, occlusion, noise, and spatial transformation by simulating developmental exposure to structured environmental variability under homeostatic constraints.
    \item Extensive experiments on benchmark datasets demonstrate that DC-CCNN++ consistently outperforms the original DC-CCNN on both classification and part segmentation tasks, while achieving stronger and more balanced robustness under various corruption settings.
\end{itemize}

\section{Related Work}
\noindent\textbf{Point Cloud Analysis Methods.}
Classical point cloud analysis methods primarily rely on geometric priors and statistical processing, including segmentation, handcrafted feature extraction, voxelization, and graph construction~\cite{li2022comprehensive,gonizzi2024evaluation}. While such methods are often computationally efficient and relatively interpretable, their effectiveness is highly sensitive to noise, missing points, and irregular sampling density. More importantly, they generally lack the hierarchical representation capacity required for large-scale and semantically rich 3D understanding tasks.
With the development of deep learning, point cloud analysis has gradually evolved into several major paradigms according to data representation. Indirect methods project point clouds onto regular structures, such as voxels, multi-view images, or range maps, thereby enabling the reuse of mature 2D or 3D convolutional backbones. Although effective in certain scenarios, these projection-based strategies inevitably introduce quantization errors or geometric information loss, particularly under sparse or incomplete observations. In contrast, direct point-based methods operate on raw unordered point sets and avoid handcrafted discretization. PointNet~\cite{qi2017pointnet} first demonstrated the feasibility of permutation-invariant point feature learning through shared MLPs and symmetric aggregation, whereas PointNet++~\cite{qi2017pointnet++} further introduced hierarchical local grouping to capture multi-scale geometric structures. Subsequent point-based architectures, such as PointMLP~\cite{ma2022rethinking}, further indicate that competitive performance can still be achieved without resorting to overly complex convolutional or attention-based operators, provided that local structural modeling is appropriately designed.

Despite these advances, robustness remains a fundamental challenge in point cloud analysis. Most existing methods are developed and evaluated under relatively clean benchmark settings, whereas real-world 3D perception systems must operate under sparse sampling, local occlusion, outliers, sensor noise, and spatial transformations. Recent robustness-oriented studies~\cite{ren2022benchmarking,yan2024benchmarking} have shown that point cloud models may suffer severe degradation under corruption, suggesting that strong clean-data performance does not necessarily translate into reliable behavior in practical environments. This limitation highlights the need for architectures that can not only extract discriminative features from irregular point sets, but also maintain stable representations under structured distribution shifts.

\noindent\textbf{Convolutional Neural Networks.}
Convolutional Neural Networks (CNNs) and their variants remain among the most influential paradigms in 3D vision. Their success is largely attributable to local receptive fields, parameter sharing, and strong feature abstraction capabilities. In point cloud analysis, CNN-based frameworks are commonly realized either by regularizing the input into voxel or image representations, or by designing point-wise/local convolution operators that approximate neighborhood aggregation on irregular sets. Representative models such as PointNet~\cite{qi2017pointnet}, PointNet++~\cite{qi2017pointnet++}, and PointMLP~\cite{ma2022rethinking} have demonstrated that local feature extraction and hierarchical aggregation are essential for effective point cloud understanding. Nevertheless, even after adaptation to point-set data, many CNN-style models still inherit several limitations from conventional dense neural operators.

First, existing models often remain computationally expensive when processing large-scale or high-density point clouds, especially when repeated neighborhood search, local grouping, or dense feature projection is required. Their reliance on continuous floating-point activations and dense convolution-like operations further limits their applicability in low-power deployment scenarios. Second, standard CNN-based representations are typically optimized for average-case performance on clean data and are not explicitly designed to withstand corruption. Consequently, they may overfit fragile local evidence and exhibit unstable behavior under sparsity, occlusion, or measurement noise. Third, most CNN backbones~\cite{stearns2024curvecloudnet,qian2022pointnext,mohammadi2024point} lack explicit mechanisms for regulating feature pathways under uncertain sensory conditions. Once local evidence is corrupted, the model generally continues to rely on static feedforward transformations and a single global readout strategy, which may amplify unreliable activations or introduce bias at the decision stage.

Recent studies have therefore moved toward more efficient or more robust point cloud backbones, including lightweight MLP-style networks, Transformer variants, and corruption-aware evaluation protocols. Although these efforts have improved the trade-off among efficiency, accuracy, and robustness to some extent, they remain largely confined to the conventional deep learning paradigm. How to incorporate more biologically plausible dynamic regulation mechanisms into point cloud architectures, so as to enhance stability and adaptability under complex environmental conditions, remains insufficiently explored.

\noindent\textbf{Spiking Neural Networks.}
Spiking Neural Networks (SNNs)~\cite{yi2023learning} are biologically inspired models that emulate neural communication through discrete, asynchronous spikes~\cite{hodgkin1952quantitative}. Owing to their event-driven computation and sparse activation patterns, SNNs have attracted increasing interest as an energy-efficient alternative to conventional deep neural networks. They have shown considerable promise in sequential perception and neuromorphic computing, particularly in scenarios where the input is sparse and temporally structured~\cite{zhong2024towards,zhang2025staa,liu2025sota}. These properties make SNNs naturally appealing for point cloud analysis, since point sets are inherently sparse, irregular, and geometrically structured.

Recent efforts have begun to transfer SNN principles into 3D point cloud understanding~\cite{ren2024spiking,wu2024point,takaghaj2024efficient}. For instance, Spiking PointNet~\cite{ren2024spiking} incorporates thresholding and membrane decay into point-based processing, showing that spike-driven computation can reduce training overhead and, in certain cases, even outperform conventional deep neural counterparts. PointLCA-Net~\cite{takaghaj2024efficient} further introduces sparse competitive coding into point cloud classification. These studies suggest that neuromorphic computation can provide a viable alternative for point-based 3D analysis. However, existing SNN-based point cloud models still face several important limitations.

On the one hand, standard SNNs~\cite{hu2024lmfunet} remain simplified abstractions of biological neurons. Their dynamics are often deterministic and periodic under repetitive stimulation, which differs substantially from the stochastic and chaotic response characteristics observed in cortical neurons. This discrepancy may lead to information loss and limited representational flexibility, especially when modeling complex point cloud structures. On the other hand, current SNN-based methods for point cloud analysis primarily emphasize energy efficiency or spike sparsity, whereas their robustness under corrupted point clouds remains insufficiently investigated. In particular, they generally lack system-level mechanisms for stabilizing feature pathways and training strategies capable of adapting to continuously changing sensory statistics. Therefore, although SNNs offer a compelling direction for efficient point cloud analysis, how to preserve their energy-efficient characteristics while improving robustness under complex perturbations remains an open and important problem.


\begin{figure*}[t!]
\centering
\includegraphics[width=0.95\linewidth]{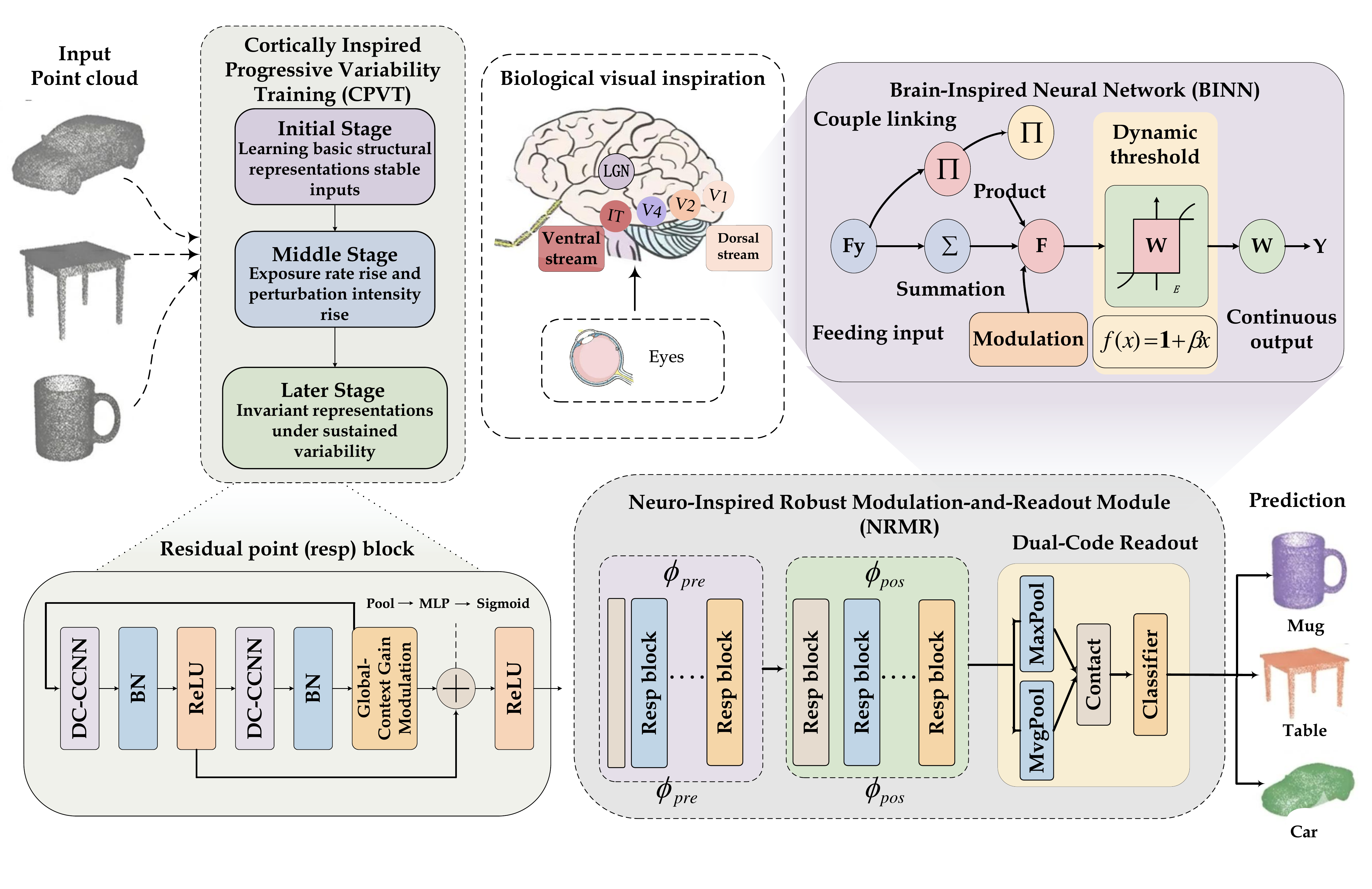}
\caption{Pipeline overview of DC-CCNN++ for 3D point cloud classification.Inspired by the visual cortex, the framework combines BINN-based local neural dynamics with two robustness-oriented extensions: the Neuro-Inspired Robust Modulation-and-Readout Module (NRMR) in the inference path and the Cortically Inspired Progressive Variability Training (CPVT) strategy in the training path. Input point clouds are encoded into structured representations, enhanced through global-context gain modulation and dual-code readout, and finally classified to produce robust predictions under corrupted conditions.
}
\label{fig:framework}
\end{figure*}


\section{Method}
Early point cloud models lacked robustness to noise, sparsity, and irregularity. To address these challenges, we proposed a BINN module, as illustrated in Figure~\ref{fig:framework}, which is inspired by biological neural systems. The module is sensitive to spatio-temporal signals and mimics brain-like processing to enhance performance.
Spiking PointNet~\cite{ren2024spiking} later introduced BINN concepts into point cloud analysis, incorporating SNN features such as thresholding and exponential decay to reduce energy consumption. However, SNNs remain simplified abstractions of spiking neurons, differing from biological cortical structures. Their periodic responses under repetitive input contrast with the chaotic behavior of biological neurons, leading to information loss and limited feature extraction. As a result, Spiking PointNet suffers from reduced accuracy and limited practical applicability.
To overcome these issues, we developed DC-CCNN, which introduces neuronal firing randomness and continuous outputs to mitigate information loss and better represent point cloud characteristics. Convolutional operations further improve local feature extraction, yielding a more effective processing framework.

\subsection{Spiking Neural Networks}
In complex structures like point clouds, SNNs mimic the brain's spike-based communication. Neurons receive spike trains from the previous layer and update their membrane potential, commonly modeled by the Leaky Integrate-and-Fire (LIF) equation:
\begin{equation}
\tau_m \frac{du}{dt} = -(u - u_{rest}) + R \cdot I(t), \quad u < V_{th},
\end{equation}
where $I$ is the input current, $V_{th}$ the firing threshold, and $R$, $\tau_m$ the resistance and time constant. When $u$ reaches $V_{th}$, a spike is emitted and $u$ resets to $u_{rest}$ (typically zero).

To enable training in modern frameworks (e.g., TensorFlow, PyTorch), an iterative LIF model was proposed:
\begin{align}
u_i[t + 1] &= \lambda (u_i[t] - V_{th} s_i[t]) + \sum_{j} w_{ij} s_j[t] + b_i, \\
s_i[t + 1] &= H(u_i[t + 1] - V_{th}),
\end{align}
where $w_{ij}$ is the synaptic weight, $b_i$ the bias, and $s_j[t]$ the spike from neuron $j$ at time $t$. $H(\cdot)$ is the Heaviside function, and $\lambda < 1$ is the leak factor (typically 0.2–0.25). $s_i[t]$ denotes the spike output of neuron $i$ at time $t$.

\subsection{Continuous-Coupled Neural Network}

To more accurately model the primary visual cortex and address issues such as the refractory period present in existing models, we introduce the Continuous-Coupled Neural Network (CCNN). CCNN introduces stochastic firing dynamics as opposed to the deterministic nature of Pulse-Coupled Neural Networks (PCNNs). The neuron firing state in CCNN fluctuates due to a fluctuating threshold, mimicking the chaotic dynamics observed in real primary visual cortex neurons. The membrane potential $x$ follows a Gaussian distribution for both the resting and firing states. The firing probability is computed using Bayesian inference.

The core dynamics of the CCNN model are described by the following recurrence relations:
\begin{equation}
F_{ij}(n) = e^{-a_f} F_{ij}(n-1) + V_F M_{ijkl} Y_{kl}(n-1) + S_{ij},
\end{equation}
\begin{equation}
L_{ij}(n) = e^{-a_l} L_{ij}(n-1) + V_L W_{ijkl} Y_{kl}(n-1),
\end{equation}
\begin{equation}
U_{ij}(n) = F_{ij}(n)(1 + b L_{ij}(n)),
\end{equation}
\begin{equation}
Y_{ij}(n) = \frac{1}{1 + e^{-(U_{ij}(n) - E_{ij}(n))}},
\end{equation}
where $M_{ijkl}$ and $W_{ijkl}$ are the synaptic weights, $S_{ij}$ is the external stimulus, and $b$ is the linking strength. The parameters $a_f$, $a_l$, and $a_e$ control the exponential decay factors of previous input states. The firing threshold $E_{ij}(n)$ evolves according to the following equation:
\begin{equation}
E_{ij}(n) = e^{-a_e} E_{ij}(n-1) + V_E Y_{ij}(n-1).
\end{equation}
The firing process is treated as a random fluctuation due to the hyperpolarization process, where both the resting potential and firing threshold are modeled as Gaussian-distributed random variables. The probability of firing is computed using:
\begin{equation}
P(Y_1|X) = \frac{1}{1 + e^{-(2m_1 - 2m_0)x + m_1^2 - m_0^2}{2s^2} + \ln\frac{P(Y_0)}{P(Y_1)}}.
\end{equation}

The model’s neurons exhibit periodic dynamics under constant stimuli and chaotic behavior under periodic stimuli. This is a key difference from PCNN, where the output remains periodic even with a periodic input. In contrast, CCNN shows chaotic characteristics under AC stimuli, aligning with the observed behaviors in real primary visual cortex neurons.
We propose a brain-inspired model for point cloud analysis that enhances spatiotemporal feature extraction. The framework combines a BINN module, which encodes spatial and temporal characteristics via spike-based processing, with a CCNN that models complex dynamics through iterative computation. The pipeline consists of three stages: (1) preprocessing for denoising and normalization, (2) BINN-based spatiotemporal spike encoding, and (3) CCNN-based pattern extraction and uncertainty handling. This BINN–CCNN synergy improves both dynamic modeling and spatiotemporal representation in point cloud data.

\subsection{DC-CCNN Framework}
The BINN module first processes the input  by simulating the pulse transmission mechanism of biological neural systems through its spiking neuron model. BINNs excel at handling sparse and irregular data, enabling them to capture both temporal and spatial features in the input. Particularly for dynamic and sparse data, BINNs can extract key features through spike generation and transmission. These initially extracted features serve as the foundation for subsequent processing by the CCNN module.
The CCNN module receives the feature data output from the BINN module. Its pulse convolutional layer further processes these features using a unique structure and state variables, including the input signal accumulation state F, the linking state L, and the dynamic threshold E.

\noindent\textbf{Accumulation of Output Features of F.} The input signal accumulation state F aggregates the characteristic signals output by the BINN module , simulating the continuous perception of stimulus intensity in the biological visual system and further enhancing feature representation.
\begin{equation}
F_t = \left(1 - \frac{\Delta t}{\tau_F}\right)F_{t - 1}+X.
\end{equation}
Among them, $F_t$ represents the accumulation state at time $t$; $X$ is the input signal; $\tau_F$ is the time-constant, which controls the decay rate of the input signal, $\Delta t$ represents the time step between the current and the previous time point; at the moment $t = 0$, $F=X$, and the input $X$ is the $1\times1$ feature matrix corresponding to the input of the convolutional kernel.

\noindent\textbf{The Spatial Neighborhood Coupling Effect of L.} The Linking state L models spatial neighborhood coupling by integrating the spatial relationships of the BINN output features. It simulates the effect of a pixel (neuron) being excited by its surrounding pixels, enabling the network to better capture spatial structure information.
\begin{equation}
 L_t=\left(1 - \frac{\Delta t}{\tau_L}\right)L_{t - 1}+Conv(F),
\end{equation}
where $L_t$ represents the coupling state at time $t$; $Conv(F)$ is a convolution operation that maps the accumulation state $F$ to the same number of channels as $L$; $\tau_L$ is the time-constant, which controls the decay rate of the coupling state, $\Delta t$ represents the time step between the current and the previous time point; another definition is $F_{projected}=Conv(F)$.

\noindent\textbf{Dynamic Threshold Control of E.} The dynamic threshold E regulates neuronal activation over time by adapting to accumulated input signals and local spatial interactions. This mechanism simulates homeostatic regulation in biological neurons, enabling the model to suppress background noise while enhancing sensitivity to salient patterns. The threshold rises or falls based on the difference between spike excitation U and the current threshold E, promoting adaptive filtering under varying input conditions.
\begin{equation}
E_t=\left(1 - \frac{\Delta t}{\tau_E}\right)E_{t - 1}+V_E\cdot sigmoid(U - E),
\end{equation}
where $U$ is the spike excitation, $U = F_{\text{projected}} \cdot (1 + \beta \cdot L_t)$; $\beta$ is the coupling-strength control parameter, $E_t$ is the dynamic threshold at time $t$, $\tau_E$ is the time-constant, which controls the decay rate of the dynamic threshold; $sigmoid(U - E)$ is the threshold adjustment amount; $V_E$ is the adjustment - rate control parameter.
\begin{equation}
Y = sigmoid(U - E).
\end{equation}
The sigmoid function is used to smooth the output.

\subsection{NRMR Module}
While DC-CCNN~\cite{dang2026primary} has demonstrated strong robustness by jointly modeling signal accumulation, spatial coupling, and dynamic threshold adaptation, its current design mainly operates at the level of local neuronal dynamics. Specifically, the accumulation state $F$ stabilizes temporal responses, the linking state $L$ captures neighborhood interactions, and the dynamic threshold $E$ adaptively suppresses irrelevant activations. These mechanisms enable DC-CCNN to emulate important properties of the primary visual cortex and achieve strong performance under sparse, noisy, and corrupted point cloud conditions. However, from the perspective of biological sensory systems, robust perception in complex environments is not only supported by local neuronal adaptation, but also by a higher-level system-wide adaptive mechanism that stabilizes feature pathways and integrates complementary decision cues.

In biological vision, when sensory inputs are degraded by occlusion, sparsity, or noise, the nervous system typically performs two additional operations beyond local response regulation. First, it adjusts the gain of different neural pathways according to global contextual conditions, enhancing reliable channels while suppressing unstable ones. Second, it does not rely on a single decoding statistic at the decision stage; instead, it combines complementary neural codes, such as salient local responses and population-level statistics, to obtain a more stable perceptual judgment. Inspired by this system-level adaptive behavior, we further enhance DC-CCNN with the Neuro-Inspired Robust Modulation-and-Readout Module (NRMR). Rather than introducing an independent branch, NRMR is designed as a natural extension of the original DC-CCNN framework: it supplements the existing $F$-$L$-$E$ dynamics with global pathway modulation during feature formation and dual-code evidence integration during classification.

\noindent\textbf{Global-Context Gain Modulation.}
The original DC-CCNN regulates neuronal responses through the dynamic threshold $E$, which adaptively changes according to accumulated signals and local coupling interactions. Although this mechanism is effective for local adaptive filtering, it does not explicitly model the higher-level contextual modulation that is widely observed in biological neural systems. In the visual cortex, top-down attention and neuromodulatory control often act by multiplicatively scaling the activity of different neuronal populations, thereby improving the signal-to-noise ratio of the overall representation.

To simulate this behavior, we introduce a global-context gain modulation mechanism into the residual feature extraction blocks of DC-CCNN. Given an intermediate feature map $\mathbf{X} \in \mathbb{R}^{B \times C \times L}$, we first compute a global channel descriptor by pooling each channel response:
\begin{equation}
s_c = \mathrm{Pool}(X_c), \qquad \mathbf{s} \in \mathbb{R}^{B \times C}.
\end{equation}
Based on this descriptor, a two-layer multilayer perceptron generates a channel-wise gain vector:
\begin{equation}
\mathbf{g} = \sigma(\phi(\mathbf{s})) \in (0,1)^{B \times C},
\end{equation}
where $\phi(\cdot)$ denotes the nonlinear projection implemented by the MLP, and $\sigma(\cdot)$ is the sigmoid function. The feature map is then recalibrated by multiplicative modulation:
\begin{equation}
\tilde{X}_c = g_c \odot X_c.
\end{equation}

This process can be regarded as a form of pathway-specific gain control. Different from the original threshold mechanism $E$, which mainly regulates activation at the neuron-response level, the gain vector $\mathbf{g}$ acts at the channel level and selectively reweights the reliability of entire feature pathways. In implementation, this mechanism is inserted into each residual point block after the convolutional transformation of the residual branch and before shortcut fusion.

\noindent\textbf{Dual-Code Readout by Max-Avg Pooling.}
The original classification head of DC-CCNN mainly depends on a single global pooling strategy to transform high-level features into classification logits. While this design is efficient, it may introduce bias under severe corruption, since a single readout statistic may overemphasize either local peaks or global averages, but not both. In biological decision-making, neural decoding typically integrates multiple complementary signals. Some downstream neurons are highly sensitive to the most salient evidence, while others respond according to the average activity of a population.

Motivated by this principle, we replace the single-path global readout in DC-CCNN with a dual-code readout mechanism. Given the final feature representation $\mathbf{H} \in \mathbb{R}^{B \times C \times L}$, we compute both global max pooling and global average pooling in parallel:
\begin{equation}
\mathbf{z}_{\max} = \mathrm{MaxPool}(\mathbf{H}), \qquad
\mathbf{z}_{\mathrm{avg}} = \mathrm{AvgPool}(\mathbf{H}).
\end{equation}
The two descriptors are then concatenated to form a joint readout vector:
\begin{equation}
\mathbf{z} = [\mathbf{z}_{\max}; \mathbf{z}_{\mathrm{avg}}] \in \mathbb{R}^{B \times 2C}.
\end{equation}

The max-pooled branch preserves the most salient local evidence, while the average-pooled branch captures the overall response tendency of the feature population. By combining these two descriptors, the classifier can utilize both strong local evidence and stable global statistics. Accordingly, the input dimension of the first fully connected layer is adjusted from $C$ to $2C$.

\subsection{Cortically Inspired Progressive Variability Training}

Despite the fact that DC-CCNN has improved the representation ability of the model for noisy, sparse, and irregular point clouds at the structural level through BINN--CCNN collaborative modeling, the signal accumulation state $F$, the spatial coupling state $L$, and the dynamic threshold $E$, its robustness still mainly comes from the neural dynamics design inside the network. In essence, this framework places more emphasis on how the model forms stable responses under a given input, while paying less attention to how the nervous system gradually acquires such stability under continuously changing sensory environments. From the perspective of biological visual system development, robust perception is not directly formed under clean, static, and complete inputs. Instead, during early development, the primary visual cortex is continuously exposed to low-fidelity, incomplete, and statistically fluctuating sensory stimuli, and neuronal selectivity as well as population response stability are gradually established in such persistently changing input environments. Existing neurophysiological studies have shown that cortical systems maintain stable population activity statistics through intrinsic excitability regulation and synaptic scaling, while continuously forming invariant representations across perturbation conditions by being exposed to stochastic and uncertain stimuli. Inspired by this process, we further introduce a cortically inspired progressive variability training strategy on top of DC-CCNN, which explicitly simulates controlled environmental variability during training, so that the model not only relies on structural design to withstand perturbations, but also acquires the ability to adapt to changes in input distributions through the learning process itself.

Different from conventional data augmentation, this strategy does not treat different corruption types as independent random operations, but instead unifies them as different manifestations of environmental variability during training. For point cloud inputs, environmental variability mainly appears in three aspects. The first is the change of effective support, such as information loss caused by point sparsity and local occlusion. The second is stochastic perturbation in spatial coordinates, such as the continuous measurement offset caused by Gaussian noise and the impulsive outliers introduced by salt-and-pepper noise. The third is the overall change of geometric configuration, such as spatial distribution drift caused by translation, rotation, and scaling. Although these perturbations differ in form, we argue that they all correspond in essence to the instability of input statistics that the nervous system must handle in real environments. Therefore, instead of injecting these perturbations all at once with fixed intensity and without constraint, we adopt a progressive scheduling mechanism that is closer to cortical development, so that the model can gradually adapt to increasingly complex input distributions while preserving its basic discriminative ability.

Specifically, let $T$ denote the total number of training epochs and $t$ the current training epoch. We define a variability coefficient $\alpha_t \in [0,1]$ as
\[
\alpha_t =
\begin{cases}
0, & t \leq 0.1T, \\
\dfrac{t-0.1T}{0.3T}, & 0.1T < t \leq 0.4T, \\
1, & t > 0.4T.
\end{cases}
\]
This coefficient is used to uniformly control both the application probability and the intensity of perturbations during training. Accordingly, the whole training process can be divided into three consecutive stages. At the initial stage, the model first learns basic structural representations under stable input statistics, where no perturbation or only very weak perturbation is applied, so as to avoid instability caused by excessively fluctuating input distributions before reliable representations are formed. Subsequently, as $\alpha_t$ increases linearly, both the exposure rate and the perturbation intensity gradually rise, allowing the model to move smoothly from the clean distribution to complex variability distributions. In the later stage, $\alpha_t$ remains at a high level, and the model consolidates its invariant representations under sustained variability. This training schedule, evolving from stabilization to adaptation and then to robustness, corresponds to the process by which cortical systems develop from an initially stable state to mature robust selectivity.

Considering that the nervous system does not completely lose its ability to preserve baseline stimulus statistics when adapting to changes in the external environment, we further introduce sample-level homeostatic constraints at the mini-batch level. Specifically, for the $i$-th sample in a batch, we introduce a Bernoulli mask
\[
m_i \sim \mathrm{Bernoulli}(p_t), \qquad p_t = p_{\max}\cdot \alpha_t,
\]
where $p_{\max}$ denotes the maximum variability exposure rate. When $m_i=1$, environmental variability is applied to the sample; when $m_i=0$, the sample remains as the original clean input. In this way, at any stage of training, the network always observes a mixture of perturbed and unperturbed samples. The role of this mixed-exposure mechanism is to preserve a stable statistical anchor for the training process, thereby preventing representational drift or over-adaptation to transient perturbation statistics under continuously changing input distributions. From the perspective of neuroscience, this process can be regarded as an approximation of homeostatic regulation in cortical population activity: the system continuously absorbs information from environmental variability while maintaining its basic response statistics within a controlled range.

In implementation, this training strategy keeps the input tensor dimensionality unchanged, while realizing environmental variability by modulating the effective information support and spatial statistical properties of the point cloud. For structural support perturbation, let the input point set be $\mathbf{X}\in\mathbb{R}^{N\times 3}$, and denote the effective support size by $K_t$. During training, $K_t$ is sampled from a predefined set $\mathcal{K}$, and the probability of selecting smaller support sizes gradually increases as $\alpha_t$ grows. Then $K_t$ unique points are randomly retained, and resampling is performed within the retained set to restore the original point cardinality $N$. In this way, the effective support can be progressively reduced without changing the network input interface, thereby simulating the information compression caused by point sparsity and local structural loss. For continuous coordinate perturbation, Gaussian random offsets are applied to the coordinates:
\[
\mathbf{x}'=\mathbf{x}+\epsilon,\qquad \epsilon\sim\mathcal{N}(0,\sigma_t^2),
\]
where $\sigma_t=\sigma_{\max}\cdot \alpha_t$, and clipping is applied to bound the perturbation magnitude. As $\alpha_t$ increases, the strength of continuous measurement noise gradually rises, which mimics the increasing complexity of the sensory environment. For impulsive perturbations and spatial transformations, the same progressive scheduling principle is adopted, progressively increasing their application probability and perturbation range while keeping the input interface unchanged, so that the model is continuously exposed to diverse sources of environmental variability, including local outliers, partial structural loss, and global geometric drift.

From the perspective of learning dynamics, the core effect of this training strategy is to drive the model from fitting instance-level completeness to encoding stable invariants at the distribution level. For the original DC-CCNN, $F$, $L$, and $E$ already establish biologically inspired representation mechanisms for local stimulus accumulation, neighborhood coupling, and dynamic inhibition within the network; the progressive variability training strategy further supplements an adaptive dimension that is closer to sensory development, allowing the model to form robust representations under continuously changing input statistics. The former answers how the network processes complex inputs within a single forward pass, while the latter answers how the network gradually acquires such processing ability over the entire course of learning. After combining the two, the brain-inspired interpretive chain of DC-CCNN is extended from ``spike encoding--continuous coupling--dynamic threshold regulation'' to ``spike encoding--continuous coupling--dynamic threshold regulation--progressive environmental variability adaptation'', so that the robustness of the model comes not only from structural design during inference, but also from active adaptation to uncertain sensory environments during training.

\begin{table}[!t]
\caption{Classification results on ModelNet40 dataset.}
\centering

\renewcommand{\arraystretch}{1.15}
\scriptsize
\begin{tabular}{l|cc}
\toprule
\textbf{Method}  & \textbf{OA(\%)} \\
\midrule
Spiking PointNet \cite{ren2024spiking}  & 83.5 \\
RS-CNN ~\cite{liu2019relation}  & 92.9 \\
DensePoint ~\cite{liu2019densepoint}  & 93.2 \\
PointASNL ~\cite{yan2020pointasnl} & 92.9 \\
PosPool ~\cite{liu2020closer}  & 93.2 \\
Point Trans ~\cite{engel2021point}  & 92.8 \\
MLMSPT ~\cite{zhong2021point}  & 92.9 \\
PCT ~\cite{guo2021pct}  & 93.2 \\
PointMLP~\cite{ma2022rethinking}  & 93.2 \\
DC-CCNN ~\cite{dang2026primary}  & 93.5 \\
\midrule
\rowcolor{lightgray} \textbf{DC-CCNN++(ours)}  & \textbf{93.7} \\
\bottomrule
\end{tabular}
\label{tab:modelnet40}
\end{table}

\begin{table}[!t]
\caption{Classification results on ScanObjectNN dataset.}
\centering

\renewcommand{\arraystretch}{1.15}
\scriptsize
\begin{tabular}{l|ccc}
\toprule
\textbf{Method} & \textbf{mAcc(\%)} & \textbf{OA(\%)} \\
\midrule
BGA-DGCNN~\cite{uy2019revisiting} & 75.7 & 79.7 \\
BGA-PN++ ~\cite{uy2019revisiting} & 77.5 & 80.2 \\
DRNet ~\cite{qiu2021dense} & 78.0 & 80.3 \\
GBNet ~\cite{qiu2021geometric} & 77.8 & 80.5 \\
SimpleView~\cite{goyal2021revisiting} & - & 80.5 \\
PRANet ~\cite{cheng2021net} & 79.1 & 82.1 \\
MVTN ~\cite{hamdi2021mvtn} & - & 82.8 \\
PointMLP~\cite{ma2022rethinking} & 83.9 & 85.4 \\
DC-CCNN ~\cite{dang2026primary} & 84.1 & 87.3 \\
\midrule
\rowcolor{lightgray} \textbf{DC-CCNN++(ours)} & \textbf{84.4} & \textbf{87.7} \\
\bottomrule
\end{tabular}
\label{tab:scanobjectnn}
\end{table}

\section{Experiment}

We evaluate the proposed DC-CCNN++ on three standard point cloud benchmarks, including object classification on ModelNet40 and ScanObjectNN, and part segmentation on ShapeNetPart. Following the original DC-CCNN setting, we report both standard performance and robustness under corrupted inputs. In addition, we conduct ablation studies to quantify the contributions of NRMR and CPVT.

\begin{table*}[t]
    \caption{Part segmentation results on the point cloud segmentation ShapeNetPart dataset.}
    \centering
    
    \resizebox{\textwidth}{!}{
    \begin{tabular}{l|c|c|
    c c c c c c c c c c c c c c c c}
        \toprule
        \textbf{Method} & \rotatebox{90}{Cls. mIoU} & \rotatebox{90}{Inst. mIoU} & 
        \rotatebox{90}{aero} & \rotatebox{90}{bag} & \rotatebox{90}{cap} & \rotatebox{90}{car} & \rotatebox{90}{chair} & \rotatebox{90}{phone} & \rotatebox{90}{guitar} & \rotatebox{90}{knife} & \rotatebox{90}{lamp} & \rotatebox{90}{laptop} & \rotatebox{90}{motorbike} & \rotatebox{90}{mug} & \rotatebox{90}{pistol} & \rotatebox{90}{rocket} & \rotatebox{90}{skateboard} & \rotatebox{90}{table} \\
        \midrule
        PointNet++~\cite{qi2017pointnet++} & 81.9 & 85.1 & 82.4 & 79.0 & 87.7 & 77.3 & 90.8 & 71.8 & 91.0 & 85.9 & 83.7 & 95.3 & 71.6 & 94.1 & 81.3 & 58.7 & 76.4 & 82.6 \\
        Kd-Net~\cite{klokov2017escape} & - & 82.3 & 80.1 & 74.6 & 74.3 & 70.3 & 88.6 & 73.5 & 90.2 & 87.2 & 81.0 & 94.9 & 57.4 & 86.7 & 78.1 & 51.8 & 69.9 & 80.3 \\
        SO-Net~\cite{li2018so} & - & 84.9 & 82.8 & 77.8 & 88.0 & 77.3 & 90.6 & 73.5 & 90.7 & 83.9 & 82.8 & 94.8 & 69.1 & 94.2 & 80.9 & 53.1 & 72.9 & 83.0 \\
        DGCNN~\cite{wang2019dynamic} & 82.3 & 85.2 & 84.0 & 83.4 & 86.7 & 77.8 & 90.6 & 74.7 & 91.2 & 87.5 & 82.8 & 95.7 & 66.3 & 94.9 & 81.1 & 63.5 & 74.5 & 82.6 \\
        P2Sequence~\cite{liu2019point2sequence} & - & 85.2 & 82.6 & 81.8 & 87.5 & 77.3 & 90.8 & 77.1 & 91.1 & 86.9 & 83.9 & 95.7 & 70.8 & 94.6 & 79.3 & 58.1 & 75.2 & 82.8 \\
        PointASNL ~\cite{yan2020pointasnl} & - & 86.1 & 84.1 & \textbf{84.7} & 87.9 & 79.7 & 92.2 & 73.7 & 91.0 & 87.2 & 84.2 & 95.8 & 74.4 & 95.2 & 81.0 & 63.0 & 76.3 & 83.2 \\
        PCNN~\cite{atzmon2018point} & 81.8 & 85.1 & 82.4 & 80.1 & 85.5 & 79.5 & 90.8 & 73.2 & 91.3 & 86.0 & \textbf{85.0} & 95.7 & 73.2 & 94.8 & 83.3 & 51.0 & 75.0 & 81.8 \\
        SpiderCNN ~\cite{xu2018spidercnn} & 82.4 & 85.3 & 83.5 & 81.0 & 87.2 & 77.5 & 90.7 & 76.8 & 91.1 & 87.3 & 83.3 & 95.8 & 70.2 & 93.5 & 82.7 & 59.7 & 75.8 & 82.8 \\
        PointMLP ~\cite{ma2022rethinking} & 83.9 & 85.9 & 83.4 & 83.4 & 87.3 & 80.5 & 89.9 & \textbf{78.4} & 92.0 & 88.1 & 82.6 & 96.2 & \textbf{78.1} & 95.9 & 85.4 & 64.8 & 83.0 & \textbf{84.7} \\
        DC-CCNN~\cite{dang2026primary} & 84.6 & 86.5 & \textbf{84.3} & 83.0 & 89.4 & 78.4 & 92.2 & 77.1 & 92.5 & \textbf{88.6} & 82.6 & 95.3 & 76.2 & \textbf{96.3} & 85.9 & \textbf{68.6} & \textbf{83.5} & 83.2 \\
        \midrule
        \rowcolor{lightgray} \textbf{DC-CCNN++(ours)} & \textbf{85.2} & \textbf{86.9} & 84.1 & 83.5 & \textbf{89.8} & \textbf{81.1} & \textbf{92.4} & 77.9 & \textbf{92.7} & 88.2 & 83.4 & \textbf{96.3} & 77.3 & 96.1 & \textbf{86.2} & 68.4 & \textbf{83.5} & 83.6 \\
        \bottomrule
    \end{tabular}
    
    }
    \label{tab:ablation2}
\end{table*}

\begin{table*}[t]
\centering
    \caption{Classification results on the ModelNet40 dataset~\cite{wu20153d} and ScanObjectNN~\cite{uy2019revisiting} dataset with density corruption, including overall accuracy metric (\%). \textbf{Bold} represent the top-1 results.}
    \renewcommand{\arraystretch}{1.15}
    \scriptsize
    \begin{tabular}{>{\kern-0.5\tabcolsep}l|c|c|cccccccccc<{\kern-0.5\tabcolsep}}
        \toprule
        \multirow{2}{*}{\textbf{Dataset}} &
        \multirow{2}{*}{\textbf{Model}} &
        \multirow{2}{*}{\textbf{Standard}} & 
        \multicolumn{5}{c}{\textbf{Global Sparsity}}&
        \multicolumn{5}{c}{\textbf{Occlusion}}\\
        \cmidrule(lr){4-8}
        \cmidrule(lr){9-13}
        &&& 512 & 256 & 200 & 150 & 128 & 0.1 & 0.3 & 0.5 & 0.7 & 1.0 \\
        \midrule

        \multirow{3}{*}{ModelNet40~\cite{wu20153d}} 
        & PointMLP~\cite{ma2022rethinking} & 93.2 & 92.1 & 86.6 & 75.9 & 51.7 & 37.8 & 92.7 & 92.4 & 91.7 & 86.1 & 47.8 \\
        \cmidrule(lr){2-13}
        & DC-CCNN~\cite{dang2026primary} & 93.4 & \textbf{92.6} & 89.7 & 83.8 & 71.5 & 60.3 & \textbf{93.1} & \textbf{93.2} & 92.3 & 88.1 & 51.3 \\
        \cmidrule(lr){2-13}
        & \cellcolor{lightgray}\textbf{DC-CCNN++(ours)}
        & \cellcolor{lightgray}\textbf{93.7}
        & \cellcolor{lightgray}92.2
        & \cellcolor{lightgray}\textbf{92.1}
        & \cellcolor{lightgray}\textbf{91.5}
        & \cellcolor{lightgray}\textbf{90.8}
        & \cellcolor{lightgray}\textbf{89.5}
        & \cellcolor{lightgray}92.8
        & \cellcolor{lightgray}93.0
        & \cellcolor{lightgray}\textbf{92.9}
        & \cellcolor{lightgray}\textbf{92.4}
        & \cellcolor{lightgray}\textbf{60.7} \\
        \midrule

        \multirow{3}{*}{ScanObjectNN~\cite{uy2019revisiting}} 
        & PointMLP~\cite{ma2022rethinking} & 85.4 & 81.1 & 77.3 & 69.1 & 44.2 & 39.7 & 83.7 & 77.4 & 72.0 & 61.6 & 50.5 \\
        \cmidrule(lr){2-13}
        & DC-CCNN~\cite{dang2026primary} & 87.3 & 84.3 & 81.4 & 74.8 & 63.7 & 54.8 & 86.0 & 83.5 & 79.3 & 70.4 & 63.9 \\
        \cmidrule(lr){2-13}
        & \cellcolor{lightgray}\textbf{DC-CCNN++(ours)}
        & \cellcolor{lightgray}\textbf{87.7}
        & \cellcolor{lightgray}\textbf{84.9}
        & \cellcolor{lightgray}\textbf{82.7}
        & \cellcolor{lightgray}\textbf{81.3}
        & \cellcolor{lightgray}\textbf{79.8}
        & \cellcolor{lightgray}\textbf{78.5}
        & \cellcolor{lightgray}\textbf{86.5}
        & \cellcolor{lightgray}\textbf{85.2}
        & \cellcolor{lightgray}\textbf{83.1}
        & \cellcolor{lightgray}\textbf{81.3}
        & \cellcolor{lightgray}\textbf{72.8} \\
        \bottomrule
    \end{tabular}
    \label{tab:dense}
\end{table*}

\begin{table*}[h]
\centering
    \caption{Classification results on the ModelNet40 dataset~\cite{wu20153d} and ScanObjectNN dataset~\cite{uy2019revisiting} with corruption are reported, including overall accuracy (\%) for both noise and spatial corruption. \textbf{Bold} represent the top-1 results.}
    \renewcommand{\arraystretch}{1.15}
    \scriptsize
    \begin{tabular}{>{\kern-0.5\tabcolsep}l|c|c|cccccccccc<{\kern-0.5\tabcolsep}}
        \toprule
        \multirow{2}{*}{\textbf{Dataset}} &
        \multirow{2}{*}{\textbf{Model}} &
        \multirow{2}{*}{\textbf{Standard}} & 
        \multicolumn{4}{c}{\textbf{Guassion}}&
        \multicolumn{5}{c}{\textbf{Salt-and-Pepper}}&
        {\textbf{Spatial}} \\
        \cmidrule(lr){4-7}
        \cmidrule(lr){8-12}
        \cmidrule(lr){13-13}
        &&& 0.01 & 0.03 & 0.05 & 0.09 & 0.01 & 0.03 & 0.1 & 0.3 & 0.5 & Random\\
        \midrule

        \multirow{3}{*}{ModelNet40~\cite{wu20153d}} 
        & PointMLP~\cite{ma2022rethinking} & 93.2 & 88.7 & 88.4 & 88.0 & 79.7 & 92.3 & 92.0 & 89.5 & 86.6 & 49.2 & 43.7 \\
        \cmidrule(lr){2-13}
        & DC-CCNN~\cite{dang2026primary} & 93.4 & 91.9 & 90.9 & 90.7 & 90.8 & 92.7 & 92.9 & 92.4 & 91.5 & 62.8 & 71.0 \\
        \cmidrule(lr){2-13}
        & \cellcolor{lightgray}\textbf{DC-CCNN++(ours)} 
        & \cellcolor{lightgray}\textbf{93.7} 
        & \cellcolor{lightgray}\textbf{92.5} 
        & \cellcolor{lightgray}\textbf{92.5} 
        & \cellcolor{lightgray}\textbf{92.8} 
        & \cellcolor{lightgray}\textbf{92.9} 
        & \cellcolor{lightgray}\textbf{93.2} 
        & \cellcolor{lightgray}\textbf{93.4} 
        & \cellcolor{lightgray}\textbf{93.0} 
        & \cellcolor{lightgray}\textbf{91.6} 
        & \cellcolor{lightgray}\textbf{69.7} 
        & \cellcolor{lightgray}\textbf{90.1} \\
        \midrule

        \multirow{3}{*}{ScanObjectNN~\cite{uy2019revisiting}} 
        & PointMLP~\cite{ma2022rethinking} & 85.4 & 80.0 & 80.0 & 78.6 & 64.9 & 84.3 & 82.8 & 74.5 & 67.3 & 37.3 & 47.9 \\
        \cmidrule(lr){2-13}
        & DC-CCNN~\cite{dang2026primary} & 87.3 & 83.8 & 83.1 & 83.3 & 79.6 & 84.8 & 83.2 & 79.1 & 73.6 & 54.1 & 68.5 \\
        \cmidrule(lr){2-13}
        & \cellcolor{lightgray}\textbf{DC-CCNN++(ours)} 
        & \cellcolor{lightgray}\textbf{87.7} 
        & \cellcolor{lightgray}\textbf{84.3} 
        & \cellcolor{lightgray}\textbf{83.6} 
        & \cellcolor{lightgray}\textbf{84.1} 
        & \cellcolor{lightgray}\textbf{81.6} 
        & \cellcolor{lightgray}\textbf{85.7} 
        & \cellcolor{lightgray}\textbf{84.5} 
        & \cellcolor{lightgray}\textbf{82.1} 
        & \cellcolor{lightgray}\textbf{80.4} 
        & \cellcolor{lightgray}\textbf{72.3} 
        & \cellcolor{lightgray}\textbf{75.8} \\
        \bottomrule
    \end{tabular}
    \label{tab:noise_transform}
\end{table*}





\subsection{Datasets}

We evaluate our model on three standard point cloud benchmarks. ModelNet40 includes 12,311 CAD models from 40 categories, with 9,843 used for training and 2,468 for testing. ScanObjectNN comprises over 15,000 real-world object point clouds from RGB-D indoor scenes; we use the PB-T50-RS split. ShapeNetPart contains 16,881 shapes from 16 categories, each annotated with 50 fine-grained part labels, making it suitable for part segmentation.

\subsection{Implementation Details}

Our final model, denoted as DC-CCNN++, is built upon the original DC-CCNN by introducing two additional components: the Neuro-Inspired Robust Modulation-and-Readout (NRMR) module and the Cortically Inspired Progressive Variability Training (CPVT) strategy. The former operates on internal feature stabilization and decision readout, while the latter improves robustness through training-stage adaptation to environmental variability.

For the backbone architecture, we keep the original DC-CCNN design unchanged unless otherwise stated. Specifically, the BINN module is first used to encode sparse input responses, and the CCNN branch then models continuous feature dynamics through signal accumulation, spatial coupling, and dynamic threshold adaptation. On this basis, we insert channel-wise global gain modulation into the residual feature extraction blocks and replace the original single global readout with a dual-code max--avg pooling head, corresponding to the Global-Context Gain Modulation and Dual-Code Readout in NRMR.

During training, we employ CPVT to model progressive environmental variability. Rather than treating the five corruption types as isolated augmentation operations, CPVT organizes them within a unified three-stage schedule. The first stage is a stable warm-up phase with no perturbation or only very weak perturbation, which allows the network to establish clean-data representations. The second stage gradually increases both corruption exposure rate and perturbation intensity. The final stage maintains high corruption coverage while preserving a portion of clean samples in each batch. Accordingly, CPVT adopts a clean/noisy mixed exposure mechanism at the sample level to maintain a stable statistical anchor, and introduces structural support perturbation, continuous coordinate perturbation, impulsive coordinate perturbation, and spatial geometric transformation at the distribution level to encourage the model to learn cross-corruption invariances.

For classification, we follow the standard settings of ModelNet40 and ScanObjectNN. For part segmentation, we use the same training and evaluation protocol as the original DC-CCNN. Unless explicitly mentioned, all quantitative comparisons are conducted under the same data split and evaluation metrics as prior work.

\begin{table*}[t]
\centering
\caption{Ablation studies were conducted on the ModelNet40 dataset~\cite{wu20153d} with corruption, evaluating overall classification accuracy (\%) under varying levels of global sparsity and salt-and-pepper noise. \textbf{Bold} and \underline{underline} represent the top-1 and top-2 results.}
\renewcommand{\arraystretch}{1.15}
\begin{tabular}{>{\kern-0.5\tabcolsep}l|c|cccccccccc<{\kern-0.5\tabcolsep}}
    \toprule
    \multirow{2}{*}{\textbf{Model}} &
    \multirow{2}{*}{\textbf{Standard}} & 
    \multicolumn{5}{c}{\textbf{Global Sparsity}}&
    \multicolumn{5}{c}{\textbf{Salt-and-Pepper}}\\
    \cmidrule(lr){3-7}
    \cmidrule(lr){8-12}
    && 512 & 256 & 200 & 150 & 128 & 0.02 & 0.03 & 0.1 & 0.3 & 0.5\\
    \midrule
    PointMLP~\cite{ma2022rethinking} & 93.2 & 92.1 & 86.6 & 75.9 & 51.7 & 37.8 & 92.3 & 92.0 & 89.5 & 86.6 & 49.2\\
    DC-CCNN ~\cite{dang2026primary} & \underline{93.4} & \underline{92.6} & \underline{89.7} & 83.8 & 71.7 & 60.3 & \underline{92.7} & \underline{92.9} & \underline{92.4} & \underline{91.5} & 62.8\\
    \midrule
    \rowcolor{lightgray} DC-CCNN++(\textbf{Global-Context}) & \textbf{93.7} & \underline{92.6} & 87.9 & 82.8 & 72.8 & 62.9 & 91.9 & 91.4 & 92.3 & 90.2 & 63.7\\
    \rowcolor{lightgray} DC-CCNN++(\textbf{Dual-Code}) & 93.6 & 92.3 & 89.4 & 84.8 & 74.1 & 62.7 & 90.4 & 90.1 & 91.7 & 89.5 & 65.9\\
    \rowcolor{lightgray} DC-CCNN++(\textbf{CPVT}) & \underline{93.4} & \textbf{92.7} & \textbf{92.3} & \underline{91.6} & \textbf{91.2} & \textbf{90.6} & \underline{91.9} & \underline{92.9} & \underline{92.9} & \underline{92.2} & \textbf{87.3}\\
    \rowcolor{lightgray} DC-CCNN++ & \textbf{93.7} & 92.2 & \underline{92.1} & \textbf{91.5} & \underline{90.8} & \underline{89.5} & \textbf{93.2} & \textbf{93.4} & \textbf{93.0} & \textbf{91.6} & \underline{69.7}\\
    \bottomrule
\end{tabular}
\label{tab:ablation}
\end{table*}


\setlength{\tabcolsep}{1pt}
\begin{table}[ht]
\centering
\caption{Ablation studies conducted on the ShapeNetPart dataset for part segmentation tasks. Results are reported using class-wise mIoU (Cls.mIoU) and instance-wise mIoU (Inst.mIoU), both in percentage (\%). Ablation studies on the ScanObjectNN dataset, reported by overall accuracy (OA).}
\renewcommand{\arraystretch}{0.3}
\footnotesize
\begin{tabular}{lccc}
\toprule
\textbf{Method}  & \textbf{Cls.mIoU(\%)} & \textbf{Inst.mIoU(\%)}& \textbf{OA(\%)} \\
\midrule
Signal Accumulation & 84.9 & 85.7 & 86.4 \\
Spatial Coupling    & 83.7 & 86.4 & 86.3 \\
Dynamic Threshold   & 83.9 & 85.9 & 85.2 \\
DC-CCNN ~\cite{dang2026primary}     & 84.6 & 86.5 & 87.3 \\
\midrule
\rowcolor{lightgray} \textbf{Global-Context}     & \textbf{85.4} & 85.9 & 86.3 \\
\rowcolor{lightgray} \textbf{Dual-Code} & 84.1 & 86.6 & 87.1 \\
\rowcolor{lightgray} \textbf{CPVT}     & 84.4 & 86.3 & 86.6 \\
\rowcolor{lightgray} \textbf{DC-CCNN++}    & 85.2 & \textbf{86.9} & \textbf{87.7} \\
\bottomrule
\end{tabular}
\label{tab:scanobjectnn_ablation}
\end{table}

\subsection{Quantitative Analysis}

Table~\ref{tab:modelnet40} reports the classification results on ModelNet40. DC-CCNN++ achieves 93.7\% OA, improving over the original DC-CCNN (93.5\%) and PointMLP (93.2\%). This indicates that the combination of NRMR and CPVT further improves representation quality while preserving the clean-data discriminative ability of the original DC-CCNN backbone.
Table~\ref{tab:scanobjectnn} presents the results on ScanObjectNN. DC-CCNN++ reaches 84.4\% mAcc and 87.7\% OA, outperforming the original DC-CCNN by 0.3\% and 0.4\%, respectively. Since ScanObjectNN contains realistic occlusion, clutter, and sensor noise, this gain shows that the proposed model generalizes better under complex real-world distributions, and also suggests that the feature modulation and robust readout of NRMR are consistent with the training-stage environmental adaptation induced by CPVT.
For part segmentation, the results on ShapeNetPart are shown in Table~\ref{tab:ablation2}. DC-CCNN++ obtains 85.2\% Cls.mIoU and 86.9\% Inst.mIoU, compared with 84.6\% and 86.5\% for the original DC-CCNN. This demonstrates that the proposed modifications are beneficial not only for object-level classification, but also for fine-grained geometric parsing.

\subsection{Robustness Studies}

We follow the original corruption protocol and evaluate the model under density corruption, noise corruption, and spatial transformation corruption. The detailed results are reported in Table~\ref{tab:dense} and Table~\ref{tab:noise_transform}. The evaluated corruption types include global sparsity, occlusion, Gaussian noise, salt-and-pepper noise, and spatial transformation.

Under global sparsity, DC-CCNN++ shows a clear advantage over both PointMLP and the original DC-CCNN, especially in the low-point regime. On ModelNet40, when the input is reduced to 128 points, DC-CCNN++ achieves 89.5\% accuracy, compared with 60.3\% for DC-CCNN. Similar improvements are observed on ScanObjectNN. This result indicates that the progressive exposure to structural support perturbation in CPVT substantially improves the model's adaptation to support-set reduction, while the representation- and readout-level stabilization provided by NRMR further amplifies this gain.

Under occlusion, DC-CCNN++ remains consistently strong across all corruption levels and yields the largest gains at severe occlusion ratios. On ModelNet40, the accuracy under the strongest occlusion rises from 51.3\% to 60.7\%, and on ScanObjectNN it increases from 63.9\% to 72.8\%. This suggests that the proposed model better preserves structural discrimination when local regions are missing, where CPVT contributes the primary adaptation to local structural damage and NRMR further stabilizes feature pathways and final decisions under missing-support conditions.

Under Gaussian noise and salt-and-pepper noise, DC-CCNN++ also achieves the best results across almost all noise levels. On ModelNet40, it improves the accuracy under Gaussian std = 0.09 from 90.8\% to 92.9\%, and under 50\% salt-and-pepper corruption from 62.8\% to 69.7\%. Similar gains are observed on ScanObjectNN under both continuous and impulsive perturbations. These results verify the explicit synergy between the training-stage adaptation of CPVT to continuous and impulsive perturbation statistics and the stabilization effect of NRMR on noise-corrupted feature pathways.

For spatial transformation, DC-CCNN++ obtains 90.1\% on ModelNet40 and 75.8\% on ScanObjectNN, significantly outperforming the original DC-CCNN. This indicates that the translation, rotation, and scaling perturbations introduced by CPVT effectively strengthen geometric invariance, while NRMR further alleviates the statistical shift caused by pose variation at the representation and readout levels.

\subsection{Ablation Studies}

We perform ablation studies on ModelNet40, ScanObjectNN, and ShapeNetPart to analyze the contribution of each newly introduced component. The compared variants include adding only SE-based global pathway gain modulation, only Dual-Code Readout, only CPVT, and the full DC-CCNN++ model.
On ModelNet40, Table~\ref{tab:ablation} shows that CPVT contributes the most under severe corruption. For example, under global sparsity with 128 points, the training-only variant reaches 90.6\%, far above the original DC-CCNN. Under salt-and-pepper noise with ratio 0.5, the same variant achieves 87.3\%. In contrast, SE and Dual mainly provide more balanced gains under clean and medium-corruption regimes. The full model achieves the best trade-off between standard accuracy and robustness.

On ScanObjectNN, the full DC-CCNN++ obtains 87.7\% OA, outperforming all single-factor variants. This indicates that the Global-Context Gain Modulation and Dual-Code Readout in NRMR, together with the progressive environmental variability adaptation in CPVT, act on representation modulation, decision integration, and distribution adaptation, respectively, and are clearly complementary under real-scene noise and clutter.
On ShapeNetPart, the SE variant provides the most obvious gain on Cls.mIoU among the single-factor settings, while Dual is more beneficial to Inst.mIoU. The training strategy also improves segmentation performance, but the full model still gives the best overall result, reaching 85.2\% Cls.mIoU and 86.9\% Inst.mIoU. This suggests that channel-wise recalibration, robust readout, and training-stage variability adaptation are also complementary in fine-grained geometric parsing.

These results show that SE corresponds to the Global-Context Gain Modulation in NRMR, Dual corresponds to the Dual-Code Robust Readout in NRMR, and the training strategy corresponds to the progressive environmental variability adaptation in CPVT. The three components operate on representation modulation, decision integration, and distribution adaptation, respectively, and their joint use accounts for the performance gains of DC-CCNN++ across clean, corrupted, and fine-grained task settings.

\section{Conclusion}

In this paper, we further extend the original DC-CCNN and propose DC-CCNN++, an enhanced brain-inspired architecture for point cloud analysis. By preserving the advantages of discrete--continuous collaborative encoding and further introducing a Neuro-Inspired Robust Modulation-and-Readout Module (NRMR) together with a Cortically Inspired Progressive Variability Training (CPVT) strategy, DC-CCNN++ improves representation stability and generalization from both the architectural and training perspectives. Experimental results show that DC-CCNN++ not only outperforms the original DC-CCNN on classification and part segmentation tasks, but also achieves stronger and more balanced robustness against sparsity, occlusion, Gaussian noise, salt-and-pepper noise, and spatial transformations. Overall, DC-CCNN++ provides an efficient, robust, and biologically grounded solution for point cloud analysis, and offers a promising direction for the application of brain-inspired methods in 3D vision, robotics, and autonomous systems.

\section*{Acknowledgments}
This work was supported by the National Natural Science Foundation of China (Grant No. U24B20186). This work was also supported by the Supercomputing Center of Lanzhou University.


\bibliographystyle{IEEEtran}
\bibliography{aaai2026}

\section*{Author Biography}

\begin{IEEEbiography}[{\includegraphics[width=1in,height=1.25in,clip,keepaspectratio]{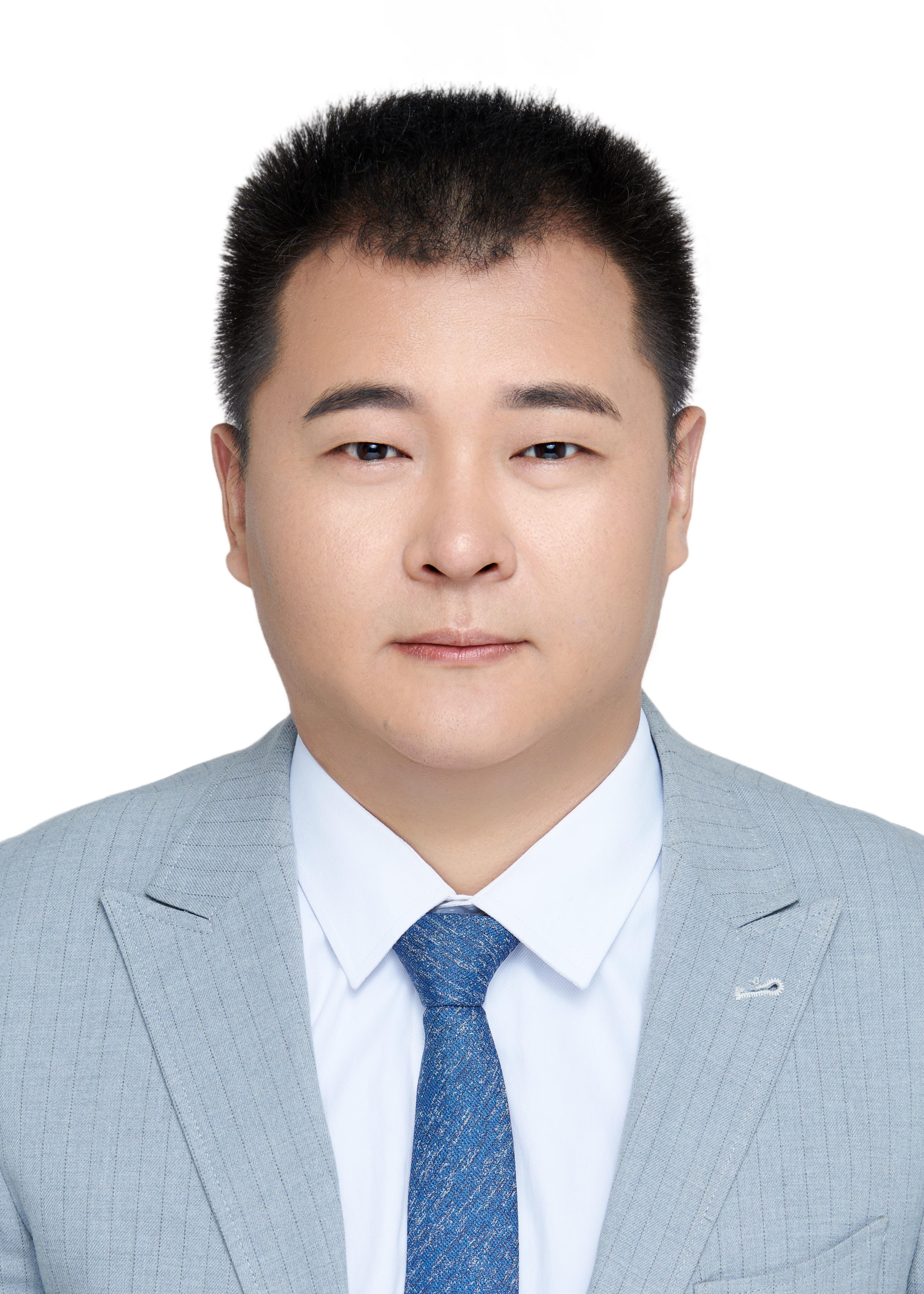}}]{Jisheng Dang}
received the Ph.D. degree from Sun Yat-sen University, China, in 2025, advised by Prof. Jianhuang Lai and Prof. Huicheng Zheng.
He worked as a research fellow at the NExT++ laboratory of the National University of Singapore, advised by Prof. Tat-Seng Chua.
He is now a tenured associate professor at the School of Information Science and Engineering, Lanzhou University.
His research interests include multimodal learning, video understanding, and embodied intelligence.
He has published several papers as the first author in major journals and conferences, including IEEE TIP, TNNLS, TITS, IJCAI, and AAAI.
He has served as a reviewer for major journals and conferences such as IEEE TPAMI, ICML, NeurIPS, ICLR, IEEE TIP, CVPR, IJCAI, ACM MM, AAAI, IEEE TMM, IEEE TCSVT, and ACM TOMM.
\end{IEEEbiography}

\begin{IEEEbiography}[{\includegraphics[width=1in,height=1.25in,clip,keepaspectratio]{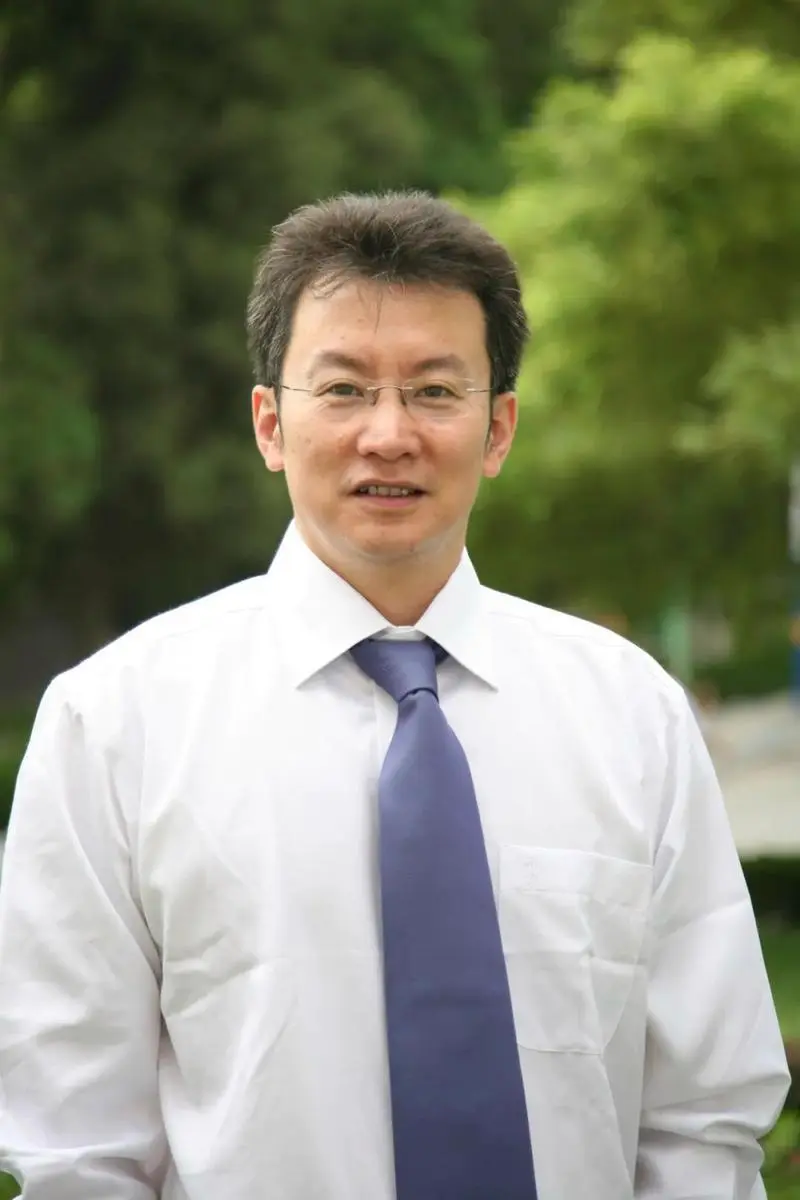}}]{Bin Hu (Fellow, IEEE)}
received the PhD degree in  computer science from the Institute of Computing Technology, Chinese Academy of Science, China, in 1998.  Since 2008, he has been a professor and the  dean of the School of Information Science and Engineering, Lanzhou University, China.  He had been also  guest professorship in ETH Zurich, Switzerland till 2011.   He is a Professor of Lanzhou University and Beijing Institute of Technology. He serves as Editor-in-Chief of IEEE Transactions on Computational Social Systems, Fellow of IET and AAIA, and Chair of Technical Committee on Computational Psychophysiology, IEEE SMC. He has published over 300 papers in domestic and international academic journals and conferences. His research interests include pervasive computing, computational psychophysiology, data  modeling, and artificial intelligence.
\end{IEEEbiography}

\begin{IEEEbiography}[{\includegraphics[width=1in,height=1.25in,clip,keepaspectratio]{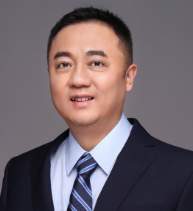}}]{Qi Tian}
(Fellow, IEEE) received the Ph.D. degree in ECE from the University of Illinois at Urbana-Champaign, Champaign, IL, USA, in 2002.
He is currently the Chief Scientist of Huawei Terminal BG.
He was the Chief Scientist in computer vision with Huawei Noah's Ark Laboratory from 2018 to 2020.
Before joining Huawei, he was a Full Professor with the Department of Computer Science, The University of Texas at San Antonio, San Antonio, TX, USA, from 2002 to 2019.
He was listed among the top ten Most Influential Scholars in Multimedia in 2016 by Aminer.org.
He is an IEEE Fellow (Class of 2016) and an Academician of the International Eurasian Academy of Sciences (elected in 2021), with 107,388 citations on Google Scholar.
\end{IEEEbiography}

\begin{IEEEbiography}[{\includegraphics[width=1in,height=1.25in,clip,keepaspectratio]{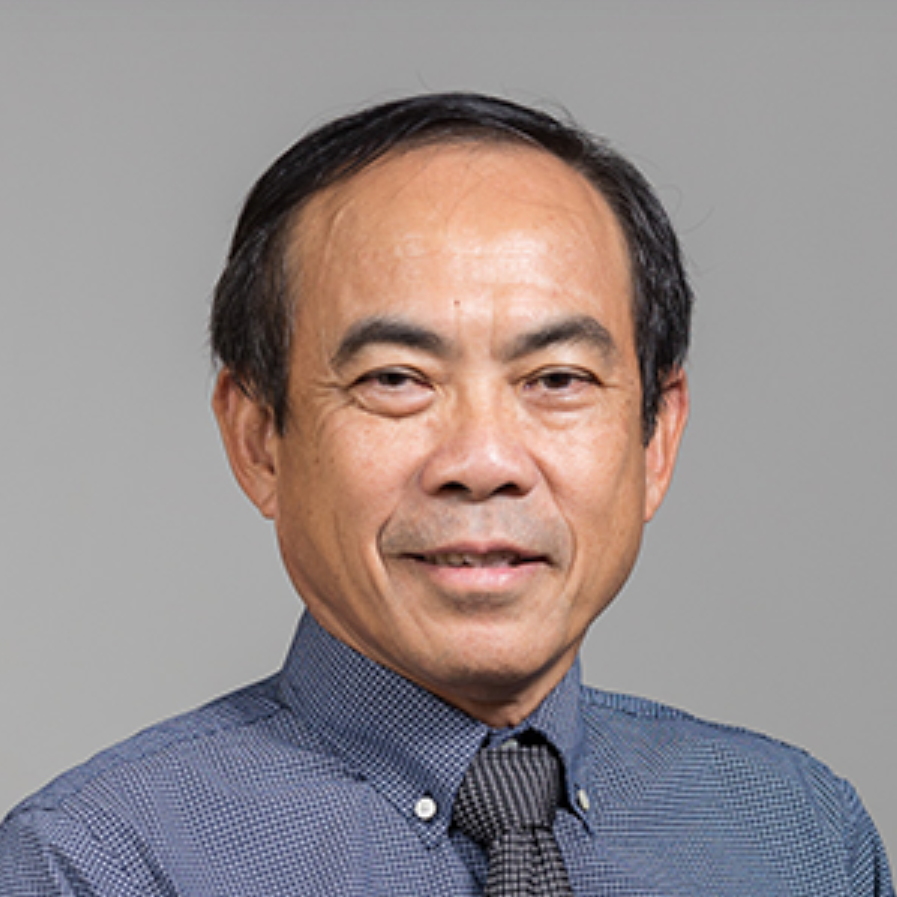}}]{Tat-Seng Chua}
received the Ph.D. degree from the University of Leeds, U.K.
He is the KITHCT Chair Professor with the School of Computing, National University of Singapore, where he was the Acting and Founding Dean of the School from 1998 to 2000.
He is the Co-Director of NExT, a joint center between NUS and Tsinghua University, to develop technologies for live social media search.
He is the 2015 winner of the prestigious ACM SIGMM Award.
He is the Chair of the Steering Committee of the ACM International Conference on Multimedia Retrieval (ICMR) and the Multimedia Modeling (MMM) conference series.
He was also the General Co-Chair of ACM Multimedia 2005, ACM CIVR (now ACM ICMR) 2005, ACM SIGIR 2008, and ACM Web Science 2015.
He serves on the editorial boards of four international journals.
He is the co-founder of two technology startups in Singapore and a Fellow of the Singapore Academy of Sciences, with     107,618 citations on Google Scholar.
\end{IEEEbiography}

\end{document}